# 基于信任系统的条件偏好协同度量框架

余 航 魏 炜 谭 征 刘惊雷
烟台大学计算机与控制工程学院 山东 烟台 264005
(yuhang8123@outlook.com)

**摘 要** 为了减少偏好度量过程中的人为干预,同时提高偏好度量算法的效率和准确性,提出一种基于信任系统的偏好协同度量框架。首先,提出了规则间的距离和规则集的内部距离等概念来具体化规则之间的关系。在此基础上,提出了基于规则集平均内部距离的规则集聚合算法 PRA,旨在保证损失最少信息的情况下筛选出最具代表性的全体用户的共同偏好,即共识偏好。之后,提出 Common belief 的概念和一种改进的信任系统,使用共识偏好作为信任系统的证据,在考虑用户一致性的同时还允许用户保留个性化信息。在信任系统下,提出了基于信任系统的有趣度度量标准,并量化了偏好的信任度和偏离度,用于描述用户偏好和信任系统的一致或相悖程度,并将用户偏好分为泛化偏好或个性化偏好,最终依据信任度和偏离度得出有趣度,从而找出最有趣的规则。在计算有趣度的过程中,提出了一个可以使用不同信任度公式来计算有趣度的可扩展的计算框架。为了进一步验证度量框架的准确性和有效性,以加权的余弦相似度公式和相关系数公式为例,提出了 IMCos 算法和 IMCov 算法。实验结果表明,信任度和偏离度有效地反映了偏好的不同特征,并且与两种最新的算法 CONTENUM 和 TKO 相比,度量框架发现的 Top-$K$ 规则在召回率、准确率和 F1-Measure 等指标上均更优。

# Contextual Preference Collaborative Measure Framework Based on Belief System

YU Hang ,WEI Wei,TAN Zheng and LIU Jing-lei
School of Computer and Control Engineering,Yantai University,Yantai,Shandong 264005,China

**Abstract** To reduce the human intervention in the preference measure process,this article proposes a preference collaborative measure framework based on an updated belief system,which is also capable of improving the accuracy and efficiency of preference measure algorithms. Firstly,the distance of rules and the average internal distance of rulesets are proposed for specifying the relationship between the rules. For discovering the most representative preferences that are common in all users,namely common preference,a algorithm based on average internal distance of ruleset,PRA algorithm,is proposed,which aims to finish the discoveryprocess with minimum information loss rate. Furthermore,the concept of Common belief is proposed to update the belief system,and the common preferences are the evidences of updated belief system. Then,under the belief system,the proposed belief degree and deviation degree are used to determine whether a rule confirms the belief system or not and classify the preference rules into two kinds(generalized or personalized),and eventually filters out Top-$K$ interesting rules relying on belief degree and deviation degree. Based on above,a scalable interestingness calculation framework that can apply various formulas is proposed for accurately calculating interestingness in different conditions. At last,IMCos algorithm and IMCov algorithm are proposed as exemplars to verify the accuracy and efficiency of the framework by using weighted cosine similarity and correlation coefficients as belief degree. In experiments,the proposed algorithms are compared to two state-of-the-art algorithms and the results show that IMCos and IMCov outperform than the other two in most aspects.

## 1 引言

近年来,随着各领域中智能设备的普及和具有上下文感知功能的数据库的发展,数据库中蕴含的信息量快速增长[1];同时,很多设备通常也配备有一些上下文传感器,如 GPS 传感器和 3D 加速度计,它们能够捕获移动用户的丰富的上下

到稿日期:2019-06-26 返修日期:2019-09-26 本文已加入开放科学计划(OSID),请扫描上方二维码获取补充信息。
基金项目:国家自然科学基金(61572419,61773331,61703360);山东省高校科学技术计划项目(J17K091)
This work was supported by the National Natural Science Foundation of China (61572419,61773331,61703360),Shandong Province University Science and technology project (J17K091).
通信作者:谭征(yttanzheng@163.com)

文信息，可以挖掘用户的个人上下文感知偏好，从而产生广泛的上下文感知服务。传统的关联规则往往仅与用户的消费行为或评价有关，忽视了产生偏好规则的上下文条件。用户偏好的不稳定性往往随着上下文的不同而变化。例如，在购物篮分析中，使用传统的关联规则方法可以获知用户会经常购买{牛奶}、{面包}、{豆浆}、{牛奶，面包}，却很少购买{豆浆，面包}，用户选择面包时更倾向于选择牛奶而不选择豆浆这一信息则被忽略。通过考虑上下文感知偏好和当前用户，可以建立个性化的上下文感知推荐系统。事实上，上下文感知推荐系统可以提供比传统推荐系统更好的用户体验[2]。但迄今为止，很少有学者致力于上下文偏好度量的研究。由于上下文偏好规则考虑了较多的信息，仅仅使用支持度和可信度很难发现哪些规则是较为有趣的；并且传统推荐系统都面临较为严重的稀疏性问题，即已知的用户偏好的数量远小于需要预测的用户偏好的数量[3]。随着上下文信息的引入，基于上下文的推荐系统将面临比前者更为严重的稀疏性问题，这进一步提升了对上下文偏好规则进行有趣度度量的难度。一方面，稀疏性问题依然存在，并且上下文偏好规则通常使用高维向量来表示；另一方面，用于度量偏好有趣度的背景知识的收集往往需要用户的参与和反馈，而用户往往无法准确地表达自己的偏好。因此，如何利用较少的信息发现准确或值得关注的上下文偏好是目前较为迫切的问题。

本文主要关注使用可扩展的偏好规则有趣度度量算法对上下文偏好规则进行有趣度度量的问题。在 Movielens 数据集上进行了一系列的研究，并使用文献[4]中的算法对数据集进行偏好挖掘，最终得出了一系列的条件偏好规则，这些规则满足支持度-可信度要求。条件偏好规则的形式为 $i^+ \succ i^- \mid X$，它表示在上下文 $X$ 的约束下，相对于后件 $i^-$，用户更偏好前件 $i^+$。在有趣度度量阶段，首先发掘了所有用户的共同偏好，并使用 PRA(Preference Rules Aggregation)算法对这些偏好进行约简；然后构建信任系统，并在信任系统下使用两种算法分别对每个用户的偏好规则进行有趣度度量。判断这些规则是否有趣主要依赖于独特性、新奇性、可靠性和简洁性[5]等性质。

以一个电影数据库为例。电影网站往往会给用户推荐一些电影信息，而这些信息并不能以很高的频率出现，否则会影响用户的体验。因此，网站往往基于用户的点击或者评分来建立每个用户的偏好数据库，从偏好数据库中发掘出有利于准确推荐的信息。同时，网站本身对提供的事务也包含一定的信息，这些信息用于描述该事务的属性以及分类，通常以上下文或标签的形式给出。一般地，一条事务被用户标有一个较高的评分，且与另一条事务有一个较大的评分差时，才认为这两条事务能构成偏好。例如，电影《头号玩家》的豆瓣评分为 8.8 分，而电影《后来的我们》的评分为 5.8 分，可以认为用户对这两部电影存在偏好。而对于《后来的我们》和《港囧》(豆瓣评分 5.6 分)这两部电影，则不能推断出用户的偏好，因为低评分的事务之间的比较是无意义的，而且这两条事务之间也不存在较大的评分差。

表 1、表 2 列出了某一用户的电影事务数据库和对应的偏好数据库。Transaction Id 表示不同的事务编号，不同的事务包含不同的属性，√ 表示该事务包含该属性。用户的偏好由两条事务构成，其中，高于 8 分的事务被认为是有较高评分的，评分之差大于 1 则被认为是存在偏好的。

表 1　事务数据库

Table 1　Transaction database

| Transaction ID | A | B | C | D | E | Rating |
|---|---|---|---|---|---|---|
| $t_1$ | √ | √ | | √ | | 9.5 |
| $t_2$ | √ | √ | | | √ | 7.4 |
| $t_3$ | √ | √ | √ | | | 6.4 |
| $t_4$ | | | √ | √ | | 8.6 |
| $t_5$ | | √ | | | √ | 7.9 |

表 2　偏好数据库

Table 2　Preference database

| Preference ID | user preference |
|---|---|
| $p_1$ | $\langle t_1, t_2 \rangle$ |
| $p_2$ | $\langle t_1, t_3 \rangle$ |
| $p_3$ | $\langle t_1, t_5 \rangle$ |
| $p_4$ | $\langle t_4, t_2 \rangle$ |
| $p_5$ | $\langle t_4, t_3 \rangle$ |

通过分析偏好 $\langle t_1, t_2 \rangle$，可以发现两条事务均包含 $A$ 和 $B$，不同的是 $t_1$ 包含 $D$ 而 $t_2$ 包含 $E$。若以这一条事务构成条件偏好规则，其语义为：用户在包含 $A$ 和 $B$ 的电影中，更喜欢包含 $D$ 的电影而不是包含 $E$ 的电影。

本文研究基于词向量表示的上下文偏好规则的有趣度度量，其特色和主要贡献如下。

(1)提出了规则之间的有向距离和规则集的平均内部距离的概念，并在此基础上设计了一种基于平均内部距离的规则集聚合算法 PRA，该算法能对规则集进行有效的低损失约简。

(2)提出了一种改进的信任系统用于偏好的协同过滤，并使用经 PRA 算法过滤后的用户的共识偏好规则作为知识背景，使得度量算法不依赖于外部给定的监督信息且具有良好的可扩展性。在此基础之上，提出了一种基于信任系统的全新的有趣度度量标准，该标准在考虑用户偏好一致性的同时，也能兼顾与共识不一致的个性化偏好。

(3)为了能在不同领域和不同特征的数据中得出准确的上下文偏好，依赖信任系统和度量标准提出了一种可扩展的有趣度度量框架。在该框架下，可灵活使用不同的计算公式计算信任度，同时也能基于信任度-偏离度-有趣度的不同特性给推荐系统提供多种偏好推荐方案。同时，为了与相关度量方法进行比较，进一步对算法的正确性和可靠性进行验证，使用加权的余弦相似度和相关系数作为有趣度计算框架中的信任度，设计了两种有趣度度量算法：基于加权的余弦相似度的算法 IMCos 和基于相关系数的算法 IMCov。

本文第 2 节介绍相关工作；第 3 节描述上下文偏好规则的相关定义；第 4 节给出信任系统、平均内部距离的定义和具体的规则集聚合算法 PRA；第 5 节介绍上下文偏好规则的有趣度度量，包括有趣度的定义、有趣度度量计算框架以及 IMCos 和 IMCov 算法；第 6 节描述有趣度度量的实验结果；最后总结全文并介绍未来的研究方向。

## 2 相关工作

有趣的上下文偏好规则的生成过程可以分为以下两个阶段。

(1)偏好规则的抽取阶段。在这一阶段中,使用关联规则是最为普遍的,也具有很高的效率[2]。现有的条件偏好挖掘算法(CP-net[3]以及以CP-net为基础的其他算法[6])能很好地处理较为复杂的偏好问题,得到的规则也较为稳定。

(2)偏好规则的度量阶段。在偏好规则度量过程中,如何使用少量的规则来代表最具价值的信息一直是备受研究者关注的问题,也是极难解决的问题。在传统的支持度-可信度框架下,改变最小支持度阈值或最小可信度阈值是较为直接有效的方法,但这取决于阈值的选择,不适当的阈值可能导致当前算法变得非常缓慢并且产生大量或太少的结果[7];同时,许多原有的框架体系对于包含上下文的偏好规则并不适用,这就要求研究者提出新的标准和方法来对关联规则的有趣度进行度量[8]。

上下文偏好生成的两个阶段的已有代表性研究工作如下。

(1)上下文偏好

文献[9]明确指出语义推理的有效性取决于其上下文,并将这一概念术语化为上下文偏好;同时还提出了一个一般化的框架,实验表明扩展了上下文的框架能明显提升模型的性能。由于用户往往会因为上下文的不同而产生不同的偏好,对此文献[10]使用层次性建模,允许上下文偏好包含多种具体的细节信息;但对于非严格的层次性结构,文中提出的方法尚无法得到很好的解决。

推荐系统中存在大量的条件偏好,而这些偏好并没有被充分考虑。文献[11]使用增强的回归树来表示推荐系统中的偏好,这种方式比目前线性的或二次函数形式的偏好更具有表现力;同时提出了一种基于梯度上升和坐标下降相结合的高效的挖掘模式。

现行的大多数算法在考虑上下文时都只单纯地考虑固定的属性值,而属性之间的交互和影响被考虑得很少。文献[12]通过提出一种属性提升(Attribute Boosting,AB)框架来进行进一步精细化建模,该框架结合概率图模型和随机梯度下降等方法的优势,使用随机采样的方法学习框架的参数,减少了盲目的随机交互带来的困扰。这一方法高效而稳定,但与上述许多算法相似,在冷启动问题上,其性能仍有所欠缺。在实际应用场景中,多上下文系统必然会面临着不一致性的问题,文献[13]详细分析了在MCS(Multi-Context System)中信息不一致问题的成因,并提出了两种偏好顺序元推理编码,扩展了MCS框架,使得无效的信息能被滤去而有效的信息能很好地被保留。

(2)偏好的有趣度

文献[14]说明了关联规则的有趣度不能简单地从支持度、可信度的角度来描述,并阐明了有趣度是一种与领域相关的概念,这表明领域可扩展算法是极为有意义的。文献[15]在给出的实例上比较分析了一些关联规则客观兴趣度度量指标,提出了使用关联规则客观兴趣度度量指标的一些建议,并给出了关联规则的有效性判定不能单纯依靠任何标准判定的结论;但实验所使用的数据集不具有代表性,结论可能存在一定偏差。

为了提升上下文偏好的准确度,文献[16]使用决策树分类算法将历史选择集合分组,在对用户进行个性化推荐时,根据上下文信息匹配最相似的分组,从而得到准确的偏好。实验表明,该方法优于传统的方法;但在匹配时仅使用余弦相似度可能会使准确性产生较大的抖动,并且在信息分类时需要大量历史信息,这会给算法带来冷启动问题。

文献[17-21]是协同过滤中的偏好度量方法。协同过滤的方法不需要指定哪些特征需要学习,也不需要大量的监督信息。这些方法先对用户进行分类,之后基于用户之间的相似性,使用相似的用户行为对其他用户进行推荐。在推荐的过程中,往往使用基于概率的方法,最为常见的有贝叶斯公式等,也有的使用低秩矩阵分解来最小化成本函数的平方误差。相关文献指出,基于协同过滤的方法不但提升了所得偏好的准确性,还可以发掘出更多的用户潜在偏好;基于用户的社交信息或行为信息,即便某些偏好不存在于用户的偏好数据库中,推荐系统也可以依据这些信息给出有效的推荐。但是,传统的协同过滤方法仍面临着冷启动和稀疏性问题。

偏好和偏好之间是具有相似性的。文献[22]从决策角度分析了偏好之间的相似性,并提出了一种基于概率的相似性度量方法,该方法使用了Kendall的tau函数。该文论证了该方法具有的稳定性和在不确定条件下的可扩展性,这对本文所提出的偏好相似和独特性理论有很大的启发。文献[23]同样是一些关于群体共识的研究,作者建立了一种2-元组语言偏序关系的群体共识优化模型和3种相应的解决方案,通过计算最小化个体一致性偏差的算术加权平均值来探讨2-元组的LPR群的偏差;该文中的结论也是极为值得关注的,即随着决策者的增加,群体共识的偏差会倾向于增加或保持不变。

本文所提出的基于信任系统的有趣度度量框架的向量化模型与已有工作主要存在以下不同。

(1)信任系统自提出后就被广泛应用在各个领域,但其概念一直没有得到较大的更新和发展。本文提出的改进信任系统Common belief基于两大经典信任系统,使得信任系统的更新更为灵活,同时兼具Hard belief和Soft belief的特征。

(2)现行的大多数基于协同过滤的有趣度度量方法仅仅考虑与知识背景具有一致性的相似偏好,而较少关注用户的个性化信息。因此,在这些度量方法中,个性化的偏好只能获得一个较低的评分。基于本文提出的改进信任系统和新的度量标准,度量框架通过使用信任度和偏离度指标能很好地将共识偏好和个性化偏好区分开并从中发掘出最有价值的偏好。

(3)大部分的现有度量算法都是用单一的有趣度计算公式或函数,而面对不同种特征的数据,有的方法并不适用,整体上缺乏一个统一而灵活的框架。本文所提出的度量框架可选取任意的计算公式作为信任度计算函数,再依据信任度得出偏离度和有趣度,极大地提升了度量算法的可扩展性。

通过上述分析可发现,虽然已有大量的工作致力于发现准确的偏好规则,但其仍然难以在实际应用场景中发挥很好

的作用。现有方法对不同应用场景下的可扩展性的考虑仍有欠缺，也没有一种很好的标准去度量偏好的有趣度，有趣的规则往往只是一些共识性的内容，而对于用户的个性化偏好特别是用户的小众偏好考虑甚少。

## 3　上下文偏好

上下文偏好规则的有趣度度量主要存在两个问题：1）有趣度标准的选择问题；2）有趣规则的选择问题。

### 3.1　上下文偏好规则的相关定义

假设 $I$ 为一个项目的集合，$X$ 是 $I$ 的子集（$X\subseteq I$），该项集对应的语言为 $L=2^I$。事务数据库 $D$ 由 $L$ 中的多项集构成，每个项集是数据库的一个元组，通常被称为事务（Transaction），形式如表 1 所列。偏好数据库 $\xi\subseteq D\times D$ 是成对事务构成的，每条记录表示事务数据库 $D$ 上用户的一条偏好。例如，对于 $\xi$ 中的一条偏好 $\langle t,u\rangle$（其中 $u$ 和 $t$ 是 $D$ 中的两条事务），其表示与 $u$ 相比用户更偏好 $t$。图 1 给出了表 1 所代表的 5 条事务和 5 条用户的偏好之间的联系。基于文献[24-25]，本文给出了上下文偏好规则及其相关定义。

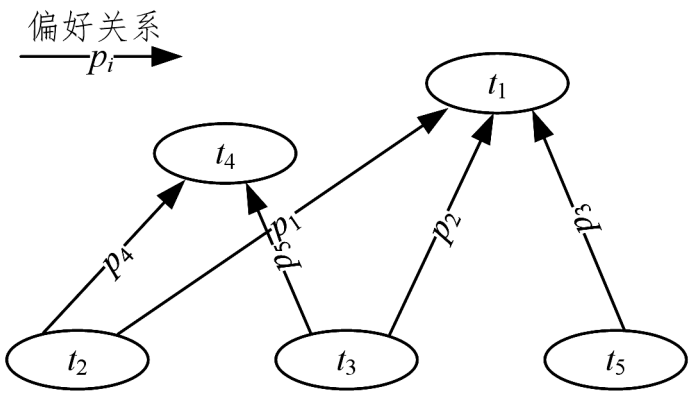

图 1　偏好关系

Fig. 1　Preference relation

**定义 1**（上下文偏好规则）　一条上下文偏好规则表示为 $i^+\succ i^-|X$，其中 $X\subseteq I$，$i^-$ 和 $i^+$ 均是 $I-X$ 中的项集。

例如，在表 1 中，一条上下文偏好规则 $D\succ E|B$ 表示在出现上下文 $B$ 时，相对于 $E$，用户更加偏好 $D$。对于偏好数据库元组事务 $t$ 和 $u$，若 $t\succ u$，且满足（$X\cup\{i^+\}\subseteq t$）$\wedge$（$X\cup\{i^-\}\subseteq u$）$\wedge$（$i^-\notin t$）$\wedge$（$i^+\notin u$），则可构成偏好规则 $R=i^+\succ i^-|X$。

**定义 2**（上下文偏好规则的支持度）　一条上下文偏好规则 $\pi$ 在偏好数据库 $\xi$ 中的支持度的定义如下：

$$supp(\pi,\xi)=\frac{agree(\pi,\xi)}{|\xi|} \tag{1}$$

支持度能反映一条上下文偏好规则被多少偏好数据库中的条目所支持。以表 1 中的 $BD$ 和 $BE$ 为例，计算其构成的上下文偏好规则 $D\succ E|B$ 的支持度：$supp(D\succ E|B,\xi)=\{p_1,p_3\}/|\xi|=0.4$。

**定义 3**（上下文偏好规则的可信度）　一条上下文偏好规则 $\pi$ 在偏好数据库 $\xi$ 中的可信度的定义如下：

$$conf(\pi,\xi)=\frac{agree(\pi,\xi)}{agree(\pi,\xi)+against(\pi,\xi)} \tag{2}$$

可信度反映的是规则与用户的偏好的一致性程度。仍以 $D\succ E|B$ 为例，其可信度为：$conf(D\succ E|B,\xi)=\{p_1,p_3\}/\{p_1,p_3\}=1.0$。

规则的支持度和规则的可信性度是一条可靠的上下文规则的基本属性，但这些属性仅仅局限于规则本身，而对于上下文偏好规则的有趣度度量是远远不够的，还需要提出一些定义来描述规则之间的关系。

**定义 4**（上下文偏好规则之间的距离）　在偏好数据库 $\xi$ 下，上下文偏好规则 $\pi_1$ 到上下文偏好规则 $\pi_2$ 的距离定义如下：

$$dis(\pi_1\rightarrow\pi_2,\xi)=P(\pi_1,\xi)-P(\pi_1\pi_2,\xi) \tag{3}$$

规则之间的距离可以用规则在偏好数据库中出现的概率来描述，但同时也与规则本身出现的概率有关。

例如，在表 1 所列的偏好数据库中，计算偏好规则 $\pi_1$：$D\succ E|B$ 到偏好规则 $\pi_2$：$D\succ B|C$ 的距离，可得 $P(\pi_1,\xi)=0.4$，$P(\pi_1\pi_2,\xi)=0$。因此，可得：$dis(\pi_1\rightarrow\pi_2,\xi)=P(\pi_1,\xi)-P(\pi_1\pi_2,\xi)=0.4-0=0.4$。

**定义 5**（事务的向量表示）　对于事务数据库 $D$ 中包含的所有属性 $attribute=\{a_1,a_2,a_3,\cdots,a_{m-1},a_m\}$，将其按某一标准唯一排序，得到一个新的序列。对于事务数据库中的一个元组，其向量表示有 $m$ 位，若第 $k$ 位为 1 则表示该事务包含属性 $a_k$，为 0 则表示 $a_k$ 不被该事务所包含。

例如，在表 1 所列的事务数据库中有 5 种不同的属性，因此事务的向量表示应有 5 位，事务 $t_1$ 的向量表示为{11010}。事务和上下文偏好规则的向量表示有利于数据的快速储存和矩阵计算。

### 3.2　上下文偏好规则的有趣度度量问题的形式化

在给定的偏好数据库下，对于已经发现的满足最小支持度和最小可信度阈值的上下文偏好规则，首要的问题是确定使用哪些标准去评价规则是否有趣。

问题 1（有趣标准选择）给定偏好数据库 $\xi$，上下文偏好规则集 $S=\{\pi_1,\pi_2,\pi_3,\cdots,\pi_{n-1},\pi_n\}$，选取 $K$ 种衡量规则有趣度的标准。

给定事务数据库的全部属性 $attribute=\{a_1,a_2,a_3,\cdots,a_{m-1},a_m\}$，上下文偏好规则最多有 $n=m\times(m-1)\times 2^{m-2}$ 条。很显然，对于较差的情况，推荐系统是无法给出准确推荐的。有趣的偏好规则的定义十分模糊，并且在实际应用场景中用户的需求不仅是不稳定的，而且往往较为多样化，因此选取多种标准是极为必要的。本文选取了文献[8]中所给出的偏好规则有趣性的 4 条性质：独特性、新奇性、可靠性、简洁性。这 4 条性质有的是相容的，有的是不相容的，其详细描述请见第 4 节。

在已经选定偏好规则有趣标准的情况下，需要解决的问题是如何利用这些已有的标准。

问题 2（有趣规则的选择）　给定偏好数据库 $\xi$、上下文偏好规则集 $S=\{\pi_1,\pi_2,\pi_3,\cdots,\pi_{n-1},\pi_n\}$ 以及 $K$ 种有趣度度量标准，对上下文偏好规则进行度量。

对于偏好规则集 $S$ 内的偏好规则，有 $K$ 种标准可以用于度量，但对于不相容的标准，偏好规则最多能表现出其中一种，如何评价一条规则是否符合一种或多种有趣度标准是较为困难的。本文建立了一个基于用户共识偏好的信任系统，在信任系统下结合信任度（belief）和偏离度（deviation），使用加权的余弦相似度和相关系数，提出了两种不同的有趣度度量算法 IMCov 和 IMCos。

## 4　信任系统

在进行偏好规则的有趣度度量的过程中，需要引入一些辅助信息。传统的主观度量过程中会要求用户提供标准化的

偏好信息，但用户往往无法准确表达这些信息。而传统的客观度量使用特定的监督信息，这会带来一些新的问题，同时也降低了算法的可扩展性。

为了降低用户的参与度并提升算法的可扩展性，本文将所有用户的共识偏好规则作为用户的一致偏好和普遍的现象纳入知识背景，以此来构建信任系统。而后，基于规则之间的有向距离提出了规则集的平均内部距离的概念，设计了一种规则集的聚合算法，用于约简共识偏好的冗余部分，减小有趣度度量过程的计算量。该算法能在不破坏规则集的特征的情况下，利用尽可能少的规则表示尽可能多的信息。

### 4.1 信任系统的定义

Silberschatz 等将信任系统分为两类[26]。

(1) Hard belief：信任系统不能被新的证据所改变。如果新的证据和信任系统发生矛盾，则认为在获得新证据的过程中产生了错误。Hard belief 是主观的，对每个用户都不一样。

(2) Soft belief：用户可以允许新证据改变这些 belief。每个 Soft belief 应分配一个等级值来表明信任度，常见的方法有贝叶斯方法、Dempster-Shafer[27] 方法和 Cyc's[28] 方法等。

以上两种信任系统分别适用于传统的主观度量和客观度量，这些信任系统的更新成本较高并且不准确，本文中所提出的信任系统 Common belief 是以上两类系统的改进。

Common belief：信任系统中的证据由全体用户给出，信任系统不可以被新的证据改变，但达到一定的阈值条件时，新的证据可以加入信任系统，而达不到阈值的证据则会被剪枝。已存在的用户行为和信任系统存在矛盾，不会改变信任系统。信任系统允许每个用户拥有与信任系统相矛盾的 Soft belief。

### 4.2 规则集的平均内部距离

规则之间的距离可以由式(3)描述，但对上下文偏好规则集仍需要进一步的定义。基于规则之间的距离，可以进一步提出规则集的平均内部距离，以描述一个规则集内规则的平均离散程度。

**定义 6**（规则集的平均内部距离） 在偏好数据库 $\xi$ 下，规则集 $S=\{\pi_1,\pi_2,\pi_3,\cdots,\pi_{n-1},\pi_n\}$ 的平均内部距离公式的描述如下：

$$avgdis(S,\xi)=\frac{\sum_{i=1}^{n}\sum_{j=1}^{n}P(\pi_i,\xi)-P(\pi_i\pi_j,\xi)}{n\times(n-1)} \tag{4}$$

其中，$n$ 是规则集 $S$ 中的规则的数目，且 $i\,!=j$。

规则集的平均内部距离是可以反映一个规则集中的规则在概率上的分布状况，规则之间同时发生的频度越低，其平均内部距离也就相应越大。

### 4.3 信任系统的建立以及规则集聚合算法 PRA

#### 4.3.1 使用共识偏好建立信任系统

若要建立一个由 Common belief 构成的信任系统用于有趣度度量，则必须先保证加入的 Common belief 是由全体用户给出的，并且不能违背大多数用户的 Soft belief（否则，这些 Common belief 就没有意义）。

借助用户的共识偏好可以很好地满足这一需求。在选取用户共识偏好建立信任系统的过程中，要确保这些共识偏好是准确的，同时还能覆盖一个较广的范围。在偏好发现过程中，首先将所有用户的偏好数据库合并，同时采用一个较高的支持度和可信度从中获取少量的偏好规则（由于数据库规模极为庞大，这些规则具有很好的可靠性和稳定性），将这些偏好规则作为信任系统的 Common belief 的候选。

信任系统、用户偏好、用户共识偏好和 Soft belief 的关系如图 2 所示。

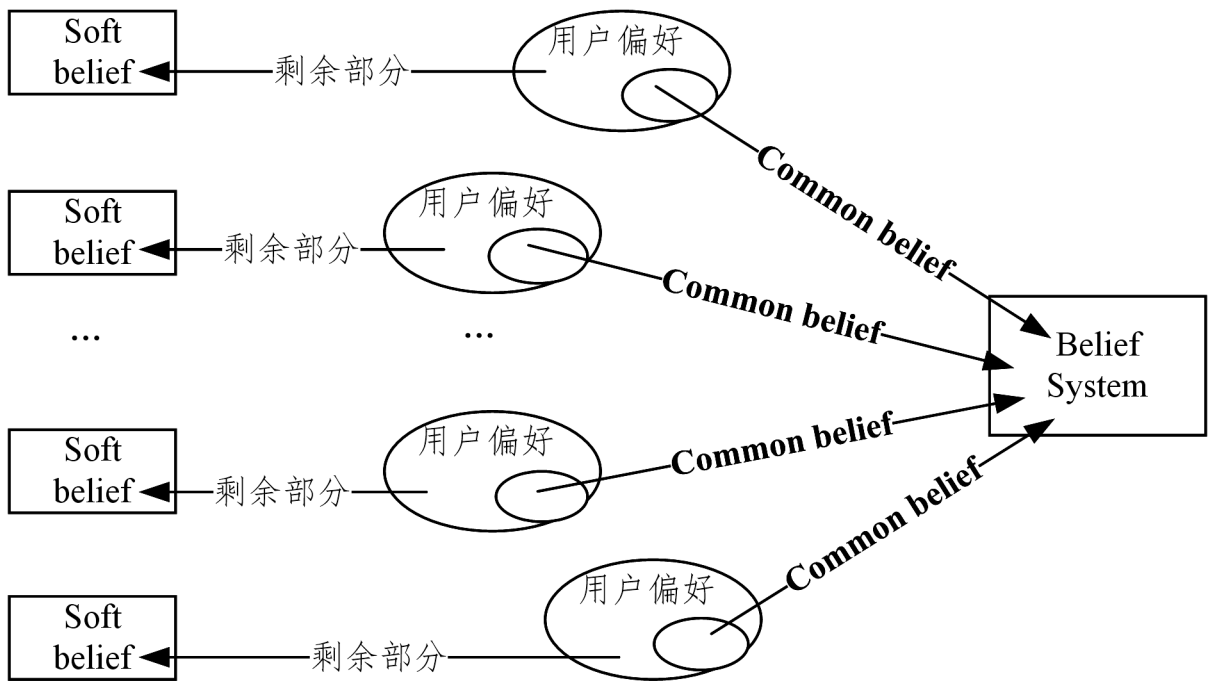

图 2 信任系统和用户偏好的关系

Fig. 2 Relationship between belief system and user preference

#### 4.3.2 偏好规则集约简算法 PRA

使用共识偏好建立信任系统会带来一个新的问题——共识偏好中往往包含着重复的信息。这些重复信息不但会影响计算的效率，还会导致偏好度量的结果被类似的偏好所影响。

为了避免这一问题，也为了得到更为有效的结果，借助规则集的平均内部距离，本文提出了一种规则集的聚合算法 PRA（见算法 1）来对偏好规则集中冗余的偏好规则进行约简。

**算法 1** 偏好规则的聚合算法 PRA

输入：偏好规则集 S，偏好数据库 ξ

输出：约简后的偏好规则集 SA

1. GIVEN Mindis
2. INITIALIZE SA = {}, Maxdis = Mindis
3. FOR ALL $\pi_i \in$ S
4. FOR ALL $\pi_j \in$ S
5. IF avgdis({$\pi_i$, $\pi_j$}, ξ) > Maxdis
6. SA = {$\pi_i$, $\pi_j$}, Maxdis = avgdis({$\pi_i$, $\pi_j$}, ξ)
7. END IF
8. END FOR
9. END FOR
10. WHILE THESE IS NEW RULE INSERTING INTO SA (SA CANNOT BE NULL)
11. Maxdis = Mindis
12. FOR ALL $\pi_i \in$ S AND $\pi_i \notin$ SA
13. IF avgdis(SA ∪ $\pi_i$, ξ) > Maxdis
14. RULE = $\pi_i$, Maxdis = avgdis(SA ∪ $\pi_i$, ξ)
15. END IF
16. END FOR
17. SA = SA ∪ RULE
18. END WHILE

算法 1 的第 1 步给定一个 *Mindis* 作为阈值，当且仅当规则集的平均内部距离大于 *Mindis* 时才允许新的规则加入。算法采用贪心的策略，首先挑选了两条距离最远的规则加入规则集，使得算法在满足阈值的条件下加入尽可能多的规则，

这一过程见第3－9行。接下来，在剩余的规则中每次挑选一条规则 $RULE$ 加入规则集 $SA$（见第10－18行），并使得规则 $RULE$ 满足 $avgdis(SA \cup RULE, \xi) > Mindis$ 且 $avgdis(SA \cup RULE, \xi) > avgdis(SA \cup \pi_i, \xi)$，其中 $\pi_i \in S, \pi_i \notin SA$（算法1的第11行、13行、14行保证了这一条件）。最后，当没有新的规则能加入到 $SA$ 中时，算法终止（见第10行）。

## 5　上下文偏好规则的有趣度度量

知识发现的核心内容之一是分辨出哪些已发现的模式是用户感兴趣的。为了达成这一目的，仅仅使用支持度-可信度框架是远远不够的。一方面，支持度-可信度框架容易受到负关联的影响；另一方面，支持度-可信度框架倾向于只描述稳定泛化的模式，用户感兴趣的内容往往不在或不限于此。因此，在已建立的信任系统下，本文提出了一种新的有趣度度量标准和计算方法，从不同的有趣度特性的角度考虑一条规则是否有趣。

### 5.1　上下文偏好规则的有趣度特性

有趣度的衡量尚没有一个统一的标准，本文在此引入几种被广泛认可的定义。

(1)独特性。在基于距离的度量下，一个模式相对于其他已发现的模式是离群的，则称这个模式是独特的。独特的模式来源于一些特殊的数据或特殊的值，这些数据或值与其余的数据和值没有较大的关联，但这些模式通常不被用户所知，因此是有趣的模式。

(2)新奇性。一个模式之前从未被发现并且不被囊括于其他任一模式中，则称这个模式具有新奇性。任一现存的数据挖掘系统都不会去发掘用户完全已知的内容，但也不能发掘出用户完全不知道的内容，因此新奇性难以以用户已知或忽略的内容去推断。

(3)简洁性。一个模式包含相对少的键值对或一组模式中含有相对少的模式时，它们都被称为具有简洁性。一个或一组具备简洁性的模式相对更容易被理解和记忆。

(4)可靠性。一个模式具有可靠性，那么它应该在应用场景中高频出现。例如若一条关联规则是可靠的，则它应该是高支持度的。一般认为，这一性质可以通过统计学、概率和信息检索等方式来进行度量。

由于这些定义之间不是完全相容的，偏好规则几乎不可能同时满足所有有趣性的性质，因此一条偏好规则被认为是有趣的，那么它至少应满足一种有趣性的性质。

### 5.2　基于信任系统的度量方法

依据信任系统发现具有简洁性和可靠性的偏好规则是较为容易的；如何使用信任系统去发现具有独特性或新奇性的规则，是接下来主要考虑的问题。

在信任系统下，一条有趣的偏好规则的有趣程度应从两方面进行描述：1)偏好规则对信任系统的信任程度；2)偏好规则对信任系统的偏离程度。

本文中，一条规则对信任系统的信任程度由该规则对信任系统内的规则的最大信任度决定，而对信任系统的偏离程度则由该规则对信任系统内的规则的平均偏离度决定。也就是说，一条有趣度的偏好规则应与信任系统中的某一条规则极为相符（可靠性的体现），或者是与信任系统中的所有规则都很不同（独特性或新奇性的体现），在满足这两种条件之一的同时，最好还能具有简洁性特征。

**定义7**（规则对信任系统的信任度）　一条偏好规则 $\pi_0$ 在数据库 $\xi$ 下对信任系统 $S=\{\pi_1,\pi_2,\pi_3,\cdots,\pi_{n-1},\pi_n\}$ 的信任度的计算公式如下：

$$belief(\pi_0 \to S, \xi) = \max(belief(\pi_0 \to \pi_i, \xi)) \quad (5)$$

其中，$i=\{1,2,3,\cdots,n\}$；$belief(\pi_0 \to \pi_i, \xi)$ 是规则之间信任度的计算公式，该公式将在4.3.1节和4.3.2节具体定义。

**定义8**（规则对信任系统的偏离度）　一条偏好规则 $\pi_0$ 在数据库 $\xi$ 下对信任系统 $S=\{\pi_1,\pi_2,\pi_3,\cdots,\pi_{n-1},\pi_n\}$ 的偏离度的计算公式如下：

$$deviation(\pi_0 \to S, \xi) = \frac{\sum_{i=1}^{n} belief(\pi_0 \to \pi_i, \xi) - 1}{n} \quad (6)$$

其中，$i=\{1,2,3,\cdots,n\}$；$belief(\pi_0 \to \pi_i, \xi)$ 是规则之间信任度的计算公式，也就是说，偏离度的定义依赖于信任度。

描述了规则对于信任系统的信任度和偏离度之后，基于偏好规则的有趣度原则，能进一步定义偏好规则的有趣度公式。

**定义9**（偏好规则的有趣度）　一条偏好规则 $\pi_0$ 在信任系统 $S=\{\pi_1,\pi_2,\pi_3,\cdots,\pi_{n-1},\pi_n\}$ 数据库 $\xi$ 中的有趣度的计算公式如下：

$$\eta(\pi_0, \xi) = \begin{cases} belief(\pi_0 \to S, \xi), \\ \quad belief(\pi_0 \to S, \xi) \geqslant abs(deviation(\pi_0 \to S, \xi)) \\ deviation(\pi_0 \to S, \xi), \\ \quad abs(deviation(\pi_0 \to S, \xi)) > belief(\pi_0 \to S, \xi) \end{cases} \quad (7)$$

本文使用余弦相似度和相关系数给出规则之间 $belief$ 的两种具体计算公式，并给出相应的两种有趣度度量算法IMCos和IMCov。

#### 5.2.1　基于余弦相似度的度量方法

在使用式(5)－式(7)计算上下文偏好规则的有趣度之前，必须先定义规则之间的信任度。

上下文偏好规则中存在着丰富的上下文，但这些上下文信息的重要程度应是不一样的，例如当上下文很长时，规则的前件和后件的重要性明显高于上下文中的一项。因此，在不进行加权的情况下，余弦相似度方法是不够准确的，本节将使用加权的余弦相似度作为规则之间的信任度。

**定义10**（基于余弦相似度的规则间信任度）　一条偏好规则 $\pi_1$ 在数据库 $\xi$ 下对另一条规则 $\pi_2$ 的基于余弦相似度的信任度计算公式如下：

$$belief(\pi_1 \to \pi_2, \xi) = \frac{k_1 \times \pi_1^+ \times \pi_2^+ + k_2 \times \pi_1^- \times \pi_2^- + k_3 \times \pi_1^X \times \pi_2^X}{\sqrt{\pi_1^+ + \pi_1^- + \pi_1^X} \times \sqrt{\pi_2^+ + \pi_2^- + \pi_2^X}} \quad (8)$$

其中，$\pi_i^+, \pi_i^-, \pi_i^X$ 是上下文偏好规则 $\pi_i(i=\{1,2\})$ 的 $i^+, i^-, X$ 的向量表示；$k_1, k_2, k_3$ 是加权系数。

根据式(5)－式(8)，给出加权的余弦相似度的偏好规则的有趣度度量算法，如算法2所示。

**算法2**　规则的有趣度度量算法IMCos

Input：信任系统S，偏好规则集RS

Output:RS 中规则的有趣度

1. GIVEN $k_1=1.2, k_2=1.2, k_3=0.6$ (WHEN $RS_i^X=0$ AND $\pi_j^X=0$, THEN $k_1=1.5, k_2=1.5, k_3=0$)
2. FOR ALL $RS_i \in$ RS
3. FOR ALL $\pi_j \in$ S
4. GET NUMBER OF 1 IN $\mathbf{RS}_i^+ \cdot \boldsymbol{\pi}_j^+$
5. GET NUMBER OF 1 IN $\mathbf{RS}_i^- \cdot \boldsymbol{\pi}_j^-$
6. GET NUMBER OF 1 IN $\mathbf{RS}_i^X \cdot \boldsymbol{\pi}_j^X$
7. GET NUMBER OF 1 IN $\mathbf{RS}_i^+ + \mathbf{RS}_i^- + \mathbf{RS}_i^X$
8. GET NUMBER OF 1 IN $\pi_j^+ + \pi_j^- + \pi_j^X$
9. GET belief($RS_i \to \pi_j, \xi$) AND deviation($RS_i \to \pi_j, \xi$)
10. END FOR
11. GET belief($RS_i \to S, \xi$) AND deviation($RS_i \to S, \xi$)
12. GET $\eta$ OF $RS_i$
13. END FOR

首先,应当给出权值系数 $k_1, k_2, k_3$ 的两种方案,分别适用于上下文为空和上下文不为空的情况(算法中给出的值为样例值而非最优值)。算法从第 2 行开始,每次选取一条用户的个性化偏好规则 $RS_i$,在第 3－10 行的过程中计算这条规则对信任系统中的每一条规则的信任度和偏离程度,并根据这一结果在第 11 行得到 $RS_i$ 与信任系统的信任度和偏离程度,最后在第 12 行依据式(7)得出规则的有趣度。

#### 5.2.2 基于相关系数的度量方法

实际上,规则之间往往并不是相互独立的,例如规则与规则可能构成关联规则中的频繁项集,而有一些规则会与另一些规则互斥。因此,本节使用相关系数来描述规则之间的信任度,以此进行有趣度度量。随着偏好数据库的增大,算法的准确性也会得到提升。

**定义 11**(基于相关系数的规则间信任度) 一条偏好规则 $\pi_1$ 在数据库 $\xi$ 下对另一条规则 $\pi_2$ 的基于相关系数的信任度的计算公式如下:

$$belief(\pi_1 \to \pi_2, \xi) = \left| \frac{Cov(\pi_1, \pi_2)}{\sigma\pi_1 \sigma\pi_2} \right| \tag{9}$$

其中,$\sigma\pi_1$ 和 $\sigma\pi_2$ 分别是 $\pi_1$ 和 $\pi_2$ 的标准差,$Cov(\pi_1, \pi_2)$ 是 $\pi_1$ 和 $\pi_2$ 的协方差。

基于式(9),可得规则的有趣度度量算法 IMCov,如算法 3 所示。

**算法 3** 规则的有趣度度量算法 IMCov

Input:信任系统 S,偏好规则集 RS

Output:RS 的有趣度

1. FOR ALL $RS_i \in$ RS
2. FOR ALL $\pi_j \in$ S
3. GET Cov($\pi_1, \pi_2$)
4. GET $\sigma\pi_1, \sigma\pi_2$
5. GET belief($RS_i \to \pi_j, \xi$) AND deviation($RS_i \to \pi_j, \xi$)
6. END FOR
7. GET belief($RS_i \to S, \xi$) AND deviation($RS_i \to S, \xi$)
8. GET $\eta$ OF $RS_i$
9. END FOR

算法 3 第 1 行从偏好规则集中选取一条规则 $RS_i$;第 2－7 行描述了在信任系统下获得这条规则对信任系统的信任度和偏离度的过程,第 3 行中求取 $Cov(\pi_1, \pi_2)$ 时可以将频率视作期望,第 4 行中计算标准差时应当使用无偏估计;规则的有趣度在算法 3 的第 8 行中依据式(7)得出。

## 6 实验与分析

本节主要通过实验对本文算法的有效性进行验证。

### 6.1 数据集和评测标准

实验采用 MovieLens-10M 和 MovieLens-20M 数据集。其中,MovieLens-10M 数据集包含在线电影推荐服务 MovieLens 的 71567 位用户应用于 10681 部电影的 10000054 个评级和 95580 个标签,所有选定的用户至少为 20 部电影进行了评分;MovieLens-20M 数据集包含了 138493 位用户对 27278 部电影的 20000263 个评级和 465564 个标签。测试集和训练集的拆分由 MovieLens 提供的脚本自动完成。

实验所使用的评价标准主要有以下几类。

(1)召回率(Recall)。在偏好数据库 $\xi$ 下,偏好规则集 $S$ 的召回率计算公式的描述如下:

$$Recall(S, \xi) = \frac{|\{\langle t, u\rangle \in \xi \mid t \succ u\}|}{|\xi|} \tag{10}$$

召回率用于描述偏好规则(集)对偏好数据库的覆盖程度,即是否较为全面地涵盖了用户的偏好。若偏好规则(集)覆盖了较多的偏好,说明这些偏好规则是泛化的,而泛化的规则往往具有可靠性。

(2)准确率(Precision)。在偏好数据库 $\xi$ 下,偏好规则集 $S$ 的准确率计算公式的描述如下:

$$Prec(S, \xi) = \frac{|\{\langle t, u\rangle \in \xi \mid t \succ u\}|}{|\{\langle t, u\rangle \in \xi, t \succ u \vee u \succ t\}|} \tag{11}$$

准确率是偏好规则(集)准确性的体现,表现规则(集)在预测偏好过程中的正确性。

(3)F1-Measure。在偏好数据库 $\xi$ 下,偏好规则集 $S$ 的 F1-Measure 计算公式的描述如下:

$$F1\text{-}Measure(S, \xi) = \frac{Prec(S, \xi) \times Recall(S, \xi)}{Prec(S, \xi) + Recall(S, \xi)} \tag{12}$$

F1-Measure 是准确率和召回率的加权平均。

(4)偏好强度(Favoritism)。在偏好数据库 $\xi$ 下,偏好规则集 $S$ 的所覆盖偏好数据库元组的平均偏好强度的计算公式的描述如下:

$$Fav(S, \xi) = \frac{|t.rating - u.rating \mid \langle t, u\rangle \in \xi, t \succ u|}{|\{\langle t, u\rangle \in \xi, t \succ u\}|} \tag{13}$$

偏好强度用于描述偏好规则(集)所覆盖的偏好的平均强度。用户对感兴趣的内容无疑会抱有强烈的偏好倾向。偏好规则覆盖的偏好的强弱程度与偏好规则的有趣程度有很大的关联。

### 6.2 PRA 算法的聚合结果

该实验的目的是验证 PRA 算法的性能,一方面是展示 PRA 算法聚合前后的共识偏好规则集,另一方面是说明通过各项指标在聚合前后的变化来表现 PRA 算法的正确性。

图 3 给出了在固定的最小可信度阈值下,使用不同的最小支持度阈值生成的共识偏好规则集在使用 PRA 算法聚合前后的召回率、准确率和偏好强度的变化。其中,PRA 算法的最小平均内部距离阈值为规则集内的规则的平均支持度的 1.5 倍,平均筛去了约 25%的规则。

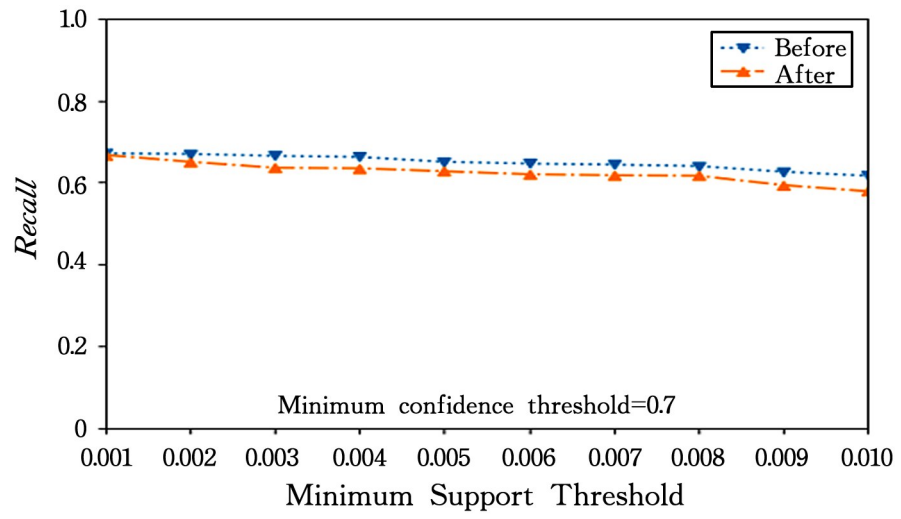

(a)召回率的变化

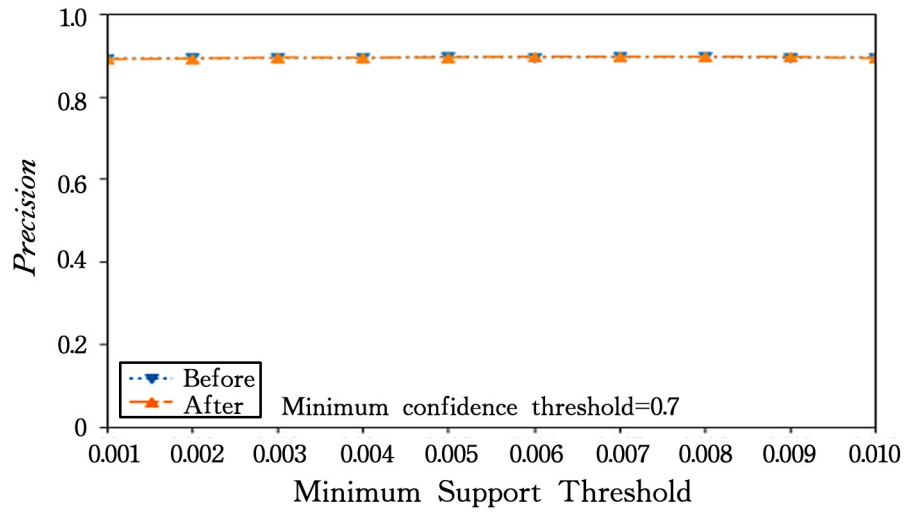

(b)准确率的变化

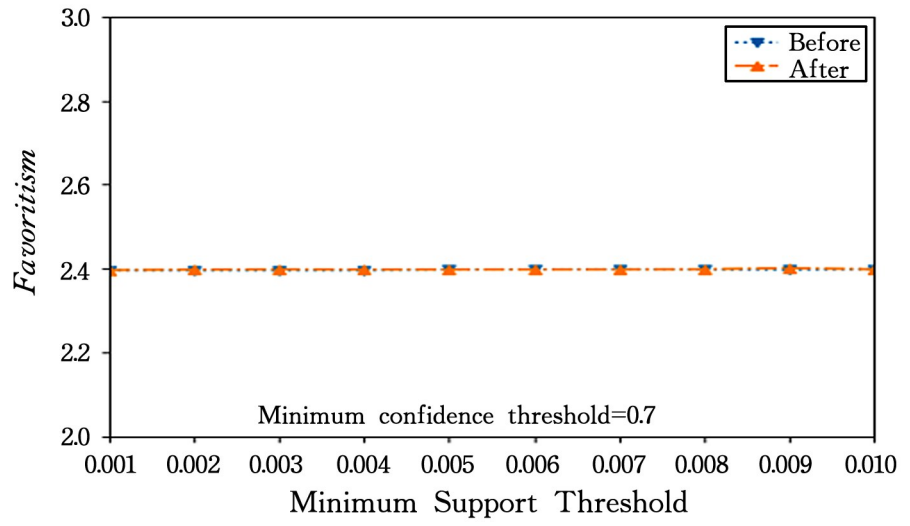

(c)偏好强度的变化

图 3　共识偏好规则集聚合前后的特征变化

Fig. 3　Feature variations of common preference rule set before and after aggregation

从图 3(a)中可以看出，共识偏好规则集的召回率是较高的，但随着最小支持度阈值的变化却没有发生较为明显的变化。这是由于生成这些偏好规则的偏好数据库包含了数万个用户，这些用户的偏好是相互独立的，一些较为泛化的规则能覆盖大部分的 Common belief，但 Soft belief 是共识偏好难以涵盖的，因此即便采用较小的最小支持度阈值也难以提高共识偏好规则集的召回率。而 PRA 算法在筛去 25％的偏好规则的同时，仅仅降低了 4％～5％的召回率，表明 PRA 算法有效降低了规则集的冗余程度，被筛去的规则大多是一些具有较高支持度的 Soft belief。在图 3(b)中，共识偏好规则集在最小可信度阈值为 0.7 的情况下生成共识偏好规则集的准确率维持在 0.9 左右，说明这些偏好规则是极为准确的，而且经过 PRA 算法约简后准确率曲线几乎没有发生变化。在图 3(c)中，偏好强度在约简后也几乎没有发生变化。

在以上 3 个实验中，知识背景规则集的准确率、召回率和偏好强度都处于较高的水平，这说明在所有用户的共识偏好中发现的偏好规则是准确且稳定的；而且通过 PRA 算法聚合后的规则集减少了 25％，其各项指标都没有发生较大变化，说明 PRA 算法很好地发挥了预期的作用，筛去了冗余的和一些个性化的偏好规则。

当最小支持度阈值为 0.01、最小可信度为 0.7 时，筛选前后的共识偏好规则集如表 3、表 4 所列。

表 3　原始的共识偏好规则集

Table 3　Original common preference rule set

| No. | $i^+$ | $i^-$ | Context | 支持度 | 可信度 |
|---|---|---|---|---|---|
| 1 | Drama | Children's | NULL | 0.0551 | 0.9710 |
| 2 | Documentary | Children's | NULL | 0.0102 | 0.9890 |
| 3 | Comedy | Children's | NULL | 0.0212 | 0.9491 |
| 4 | Romance | Children's | NULL | 0.0112 | 0.9801 |
| 5 | Crime | Children's | NULL | 0.0101 | 0.9944 |
| 6 | Thriller | Children's | NULL | 0.0148 | 0.9648 |
| 7 | War | Children's | NULL | 0.0115 | 0.9902 |
| 8 | Film-Noir | Drama | NULL | 0.0105 | 0.9893 |
| 9 | Film-Noir | Horror | NULL | 0.0163 | 1.0000 |
| 10 | Film-Noir | Comedy | NULL | 0.0142 | 1.0000 |
| 11 | Documentary | Drama | NULL | 0.0221 | 0.7778 |
| 12 | Drama | Adventure | NULL | 0.0490 | 0.7954 |
| 13 | Drama | Horror | NULL | 0.1363 | 0.9806 |
| 14 | Drama | Comedy | NULL | 0.1149 | 0.7539 |
| 15 | Drama | Musical | NULL | 0.0182 | 0.8436 |
| 16 | Crime | Drama | NULL | 0.0122 | 0.9074 |
| 17 | Drama | Action | NULL | 0.0559 | 0.8029 |
| 18 | Drama | Sci-Fi | NULL | 0.0496 | 0.8451 |
| 19 | Documentary | Horror | NULL | 0.0249 | 0.9943 |
| 20 | Documentary | Comedy | NULL | 0.0281 | 0.9064 |
| 21 | Documentary | Action | NULL | 0.0129 | 0.9137 |
| 22 | Adventure | Horror | NULL | 0.0190 | 0.9449 |
| 23 | Comedy | Horror | NULL | 0.0550 | 0.9622 |
| 24 | Mystery | Horror | NULL | 0.0145 | 1.0000 |
| 25 | Romance | Horror | NULL | 0.0249 | 0.9887 |
| 26 | Crime | Horror | NULL | 0.0249 | 0.9977 |
| 27 | Action | Horror | NULL | 0.0242 | 0.9692 |
| 28 | Thriller | Horror | NULL | 0.0312 | 0.9910 |
| 29 | War | Horror | NULL | 0.0291 | 0.9942 |
| 30 | Mystery | Comedy | NULL | 0.0124 | 0.9562 |
| 31 | Romance | Comedy | NULL | 0.0134 | 0.8166 |
| 32 | Crime | Comedy | NULL | 0.0253 | 0.9499 |
| 33 | War | Comedy | NULL | 0.0257 | 0.8844 |
| 34 | Romance | Action | NULL | 0.0102 | 0.8159 |
| 35 | Crime | Action | NULL | 0.0101 | 0.9623 |
| 36 | Thriller | Action | NULL | 0.0126 | 0.7271 |

表 4　PRA 算法聚合后的共识偏好规则集

Table 4　Common preference rule set after aggregation using PRA

| No. | $i^+$ | $i^-$ | Context | 支持度 | 可信度 |
|---|---|---|---|---|---|
| 1 | Drama | Horror | NULL | 0.1363 | 0.9806 |
| 2 | Drama | Comedy | NULL | 0.1149 | 0.7539 |
| 3 | Drama | Action | NULL | 0.0559 | 0.8029 |
| 4 | Comedy | Horror | NULL | 0.0550 | 0.9622 |
| 5 | Drama | Children's | NULL | 0.0551 | 0.9710 |
| 6 | Drama | Sci-Fi | NULL | 0.0496 | 0.8451 |
| 7 | Drama | Adventure | NULL | 0.0490 | 0.7954 |
| 8 | Documentary | Comedy | NULL | 0.0281 | 0.9064 |
| 9 | Thriller | Horror | NULL | 0.0312 | 0.9910 |
| 10 | Documentary | Horror | NULL | 0.0249 | 0.9943 |
| 11 | War | Horror | NULL | 0.0291 | 0.9942 |
| 12 | Crime | Comedy | NULL | 0.0253 | 0.9499 |
| 13 | War | Comedy | NULL | 0.0257 | 0.8844 |
| 14 | Documentary | Drama | NULL | 0.0221 | 0.7778 |
| 15 | Crime | Horror | NULL | 0.0249 | 0.9977 |
| 16 | Romance | Horror | NULL | 0.0249 | 0.9887 |
| 17 | Action | Horror | NULL | 0.0242 | 0.9692 |
| 18 | Comedy | Children's | NULL | 0.0212 | 0.9491 |
| 19 | Adventure | Horror | NULL | 0.0190 | 0.9449 |
| 20 | Drama | Musical | NULL | 0.0182 | 0.8436 |
| 21 | Film-Noir | Horror | NULL | 0.0163 | 1.0000 |
| 22 | Thriller | Children's | NULL | 0.0148 | 0.9648 |
| 23 | Film-Noir | Comedy | NULL | 0.0142 | 1.0000 |
| 24 | Mystery | Horror | NULL | 0.0145 | 1.0000 |
| 25 | Documentary | Action | NULL | 0.0129 | 0.9137 |

表 3、表 4 中，所有规则的上下文都为空。这是因为在由

大量的用户偏好元组组成的偏好数据库中，具有较长上下文的规则往往由于过于具体而很难有一个较高的支持度，这也符合一般常识。但较少的上下文信息不利于 IMCos 算法的有效度量，因此在接下来的实验中，本文减少了用户的数量(从 138493 位用户减少至 71567 位用户)，并采用了一个较低的最小支持度阈值。

### 6.3 IMCos 算法和 IMCov 算法的有趣度度量结果

该实验使用固定的最小支持度阈值 0.005 和最小可信度阈值 0.7 生成规则集，并将经由 IMCos 算法和 IMCov 算法度量后的规则集和不做处理的规则集的 Top-$K$($K$ =5,10,…,50)条规则的平均准确率、召回率和偏好强度进行比较，其中召回率和准确率的计算在测试集中完成，偏好强度的计算在训练集中完成。该实验的目的是证明 IMCos 算法和 IMCov 算法对有趣度度量的有效性，并表现信任度、偏离度和有趣度之间的关联和特性。

图 4 给出了 1000 位用户的原始偏好规则集和经 IMCos 算法度量排序后的 Top-$K$ 条偏好规则的平均召回率、准确率和偏好强度。从召回率可以看出，信任度排序的 Top-$K$ 的规则拥有较高的召回率，这是因为共识偏好规则往往是符合可靠性标准的，而一条规则对于信任系统的信任度高，说明该规则与信任系统相似，同样具有较好的可靠性，即被较多的偏好所支持。在 $K$ 值较小时，偏离度高的规则的准确率极高，远超 $K$ 值较大时的平均准确率，这是由于偏离度高的规则来自一系列或一类特殊的行为，这些行为恰恰是用户的个性化行为，因此偏离度高的规则的准确率极高。在偏好强度的实验结果中，有趣度高的偏好规则的偏好强度是最高的，说明高有趣度的规则覆盖的用户偏好都是强烈的。尽管 4 条曲线的差值不大，但偏好强度是被覆盖偏好的评分的平均值(1000 位用户的 Top-5 的规则至少覆盖了 40 万条偏好)，因此即便是极小的差值也具有十分重要的意义。

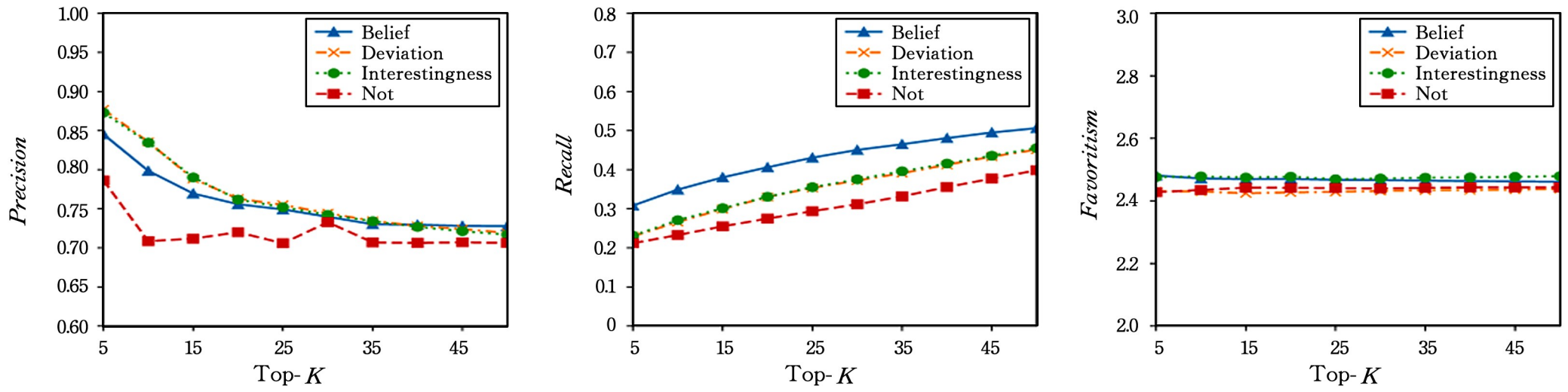

图 4 IMCos 算法的度量结果

Fig. 4 Measure results of IMCos algorithm

图 5 给出了 1000 位用户的原始偏好规则集和经 IMCov 算法度量排序后的 Top-$K$ 条的平均召回率、准确率和偏好强度。准确率最高的规则依旧是偏离度最高的规则，也正是由于这些规则来自一些用户个性化的行为，因此这些规则的召回率依旧较低(但仍然优于原始规则集)。有趣的是，召回率最高的规则不是信任度最高的规则，而是有趣度最高规则；并且在 $K$ 值较小时，召回率几乎达到了原始规则集的两倍，而且随着 $K$ 值的增大很快开始呈现收敛的趋势，这表明覆盖了较多偏好条目的规则基本已经被包括了；同时，有趣度 Top-5 的规则的平均准确率也接近 0.9，偏好强度也维持在一个相对较高的水平，这很好地体现了使用有趣度进行度量的有效性。

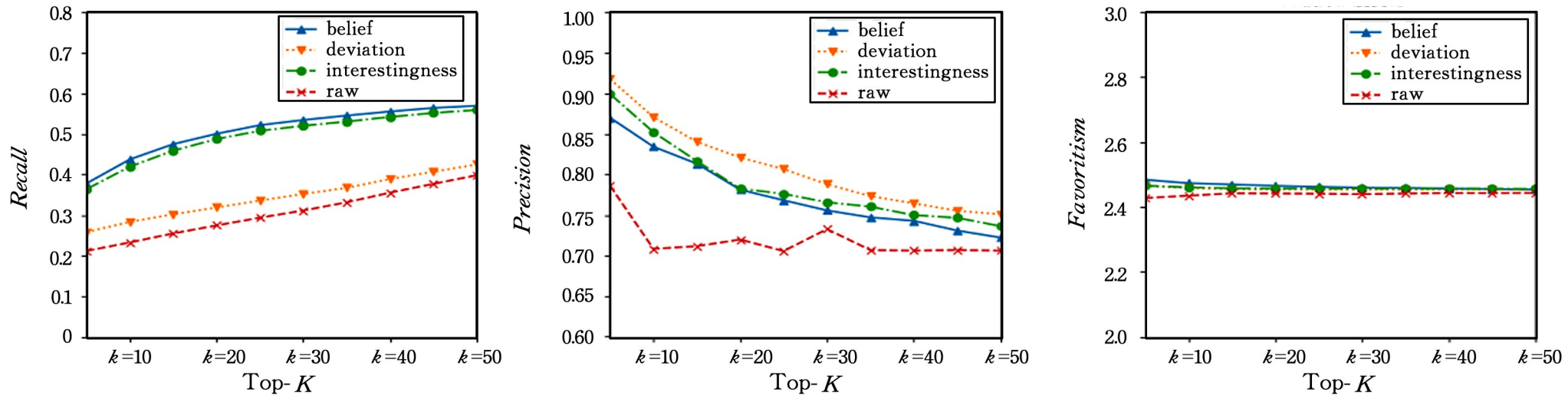

图 5 IMCov 算法的度量结果

Fig. 5 Measure results of IMCov algorithm

结合图 4 和图 5 可以得出一些值得关注的结论。首先，依照有趣度进行排序虽然并不是在所有指标上都是最优的，但在大部分情况下的结果都十分优秀，这很好地说明了有趣度是一个多方面的综合因素。此外，偏离度和信任度单独的作用也值得肯定。另一方面，在原始偏好规则集中，准确率较高或召回率较高的规则是散乱分布的，通过 IMCos 算法和 IMCov 算法均可以很好地找出这些规则，但 IMCov 算法的度量结果略优于 IMCos 算法。与此同时，IMCov 算法的开销高于 IMCos 算法(计算标准差和协方差需要较多的计算和数据库扫描)的开销，并且 IMCov 算法对偏好数据库的规模有着较为严重的依赖。因此，根据不同的应用场景选择不同的信任度计算公式是极为重要的。

综合 6.2 节和 6.3 节可以发现，具有偏离度高的规则的准确率往往都很高，但这些规则的召回率往往不高；信任度高的规则的召回率很高，但准确率相对较低。传统的观念认为，召回率和准确率是难以同时兼顾的，但使用有趣度排序时却得到了一个良好的结果。实验结果表明，基于信任系统的有趣度量方式能为推荐系统提供多种解决方案。例如，除了使用有趣度高的规则进行推荐外，还可以使用偏离度高的规则精准地推荐用户的小众偏好，或使用信任度高的规则或直接

使用共识偏好规则为新用户提供推荐。这很大程度地提升了推荐系统的性能，也一定程度地缓解了冷启动问题。

### 6.4　相关度量算法的比较

为了进一步体现度量框架的优越性，本节将从召回率、准确率和 F1-Measure 的角度，来展示 IMCos 算法和 IMCov 算法与最新的 CONTENUM[2] 算法和 TKO[29] 算法所发现 Top-$K$ 规则的对比结果，其中 $K=\{10,20,30,40,50\}$，RAW 表示未经排序的结果。

图 6(a)给出了召回率的对比结果。可以发现，IMCov 算法所发现的 Top-$K$ 规则在所有 $K$ 值的情况下所具有的召回率均明显优于其他结果。这说明在偏好数据库较大而上下文信息较为匮乏的情况下，相比基于上下文匹配的方式，基于概率的模型更为有效，并且基于信任度和偏离度的选择机制在保证准确性的同时也保障了规则具有良好的泛化性。CONTENUM 算法基于信任度和可信度对规则进行排序，表现仅次于 IMCov 算法，这表明经典的支持度可信度在有趣度范畴内仍具有良好的参考价值。尽管共识偏好中严重缺乏上下文，但基于信任度-偏离度框架的 IMCos 算法的表现依然较为良好，略优于 TKO 算法。尽管在 $K$ 值较小时由于数据库中的上下文信息不够丰富，IMCos 算法的表现稍逊于 CONTENUM 算法，但随着 $K$ 值的增大，这一差距逐渐缩小。

图 6(b)给出了各算法的准确率。可以看出，4 种算法所发现 Top-$K$ 规则的准确率均随 $K$ 值的增大而降低，并最终收敛于平均准确率。其中，IMCov 算法仍是表现最好的算法，在 $K=10$ 时，其准确率相比未经筛选的结果高出了 17%，比其他算法高出了 5%～10%。在 $K$ 值较小时，IMCos 算法仅优于 TKO 算法，而随着 $K$ 值的增大，IMCos 算法的表现趋于优秀，在 $K=50$ 时其表现仅次于 IMCov。

图 6(c)给出了 4 种算法的 *F*1-*Measure* 结果。可以很明显地得出，在这一指标下，IMCov 算法最为优异，这也符合之前的实验结果。有趣的是，随着 $K$ 值的增大，所有算法的 *F*1-*Measure* 值都有所上升，这是因为 $K$ 值的增大使得召回率逐渐上升，尽管准确率有所下降，但是所有规则的准确率都处于一个较高水平，因此产生了这一结果。RAW 数据的 *F*1-*Measure* 值呈线性增长，这也从侧面体现了 F1-Measure 值随 $K$ 值而增大。

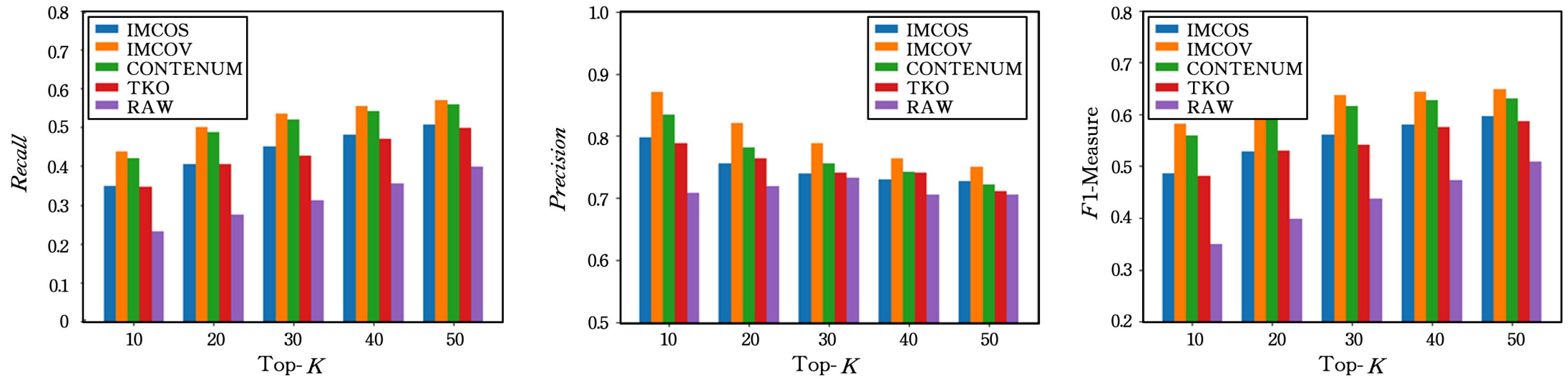

(a)召回率的比较　　(b)准确率的比较　　(c)*F*1-*Mearsure* 的比较

图 6　相关算法的比较

Fig. 6　Comparsion among relative algorithms

综合来看，IMCov 算法的表现最为优秀，在几乎所有情况下，都优于其他算法；而 IMCos 算法的表现随着数据库中上下文信息的丰富程度的提升而提升，在知识背景规则集的上下文信息为空的情况下，其表现仍优于 TKO 算法，且在 $K$ 值较大时 F1-Measure 值更为稳定。

**结束语**　本文主要对信任系统下对上下文偏好规则进行有趣度度量的问题展开了研究，定义了上下文偏好，分析了上下文偏好有趣度度量过程中所存在的问题，定性、定量地给出了上下文偏好规则的有趣度计算方法。首先，提出了一种 Common belief 的概念，并设计了一种改进的信任系统，同时给出了共识偏好聚合算法，用于进一步优化和精简信任系统，以提升其效率和准确性；在此之上，提出了一种基于信任系统的有趣度度量框架，基于信任系统计算规则对信任系统的信任度和偏离度，以此来确定一条规则是否有趣；为了对框架进行有效性验证，分别使用余弦相似度和相关系数作为范例，给出了具体的信任度、偏离度和有趣度的计算公式，给出了相应的有趣度度量算法，并相应地将整个框架向量化，以提升效率；最后，将算法 IMCos 和 IMCov 与相关的度量算法进行比较。实验结果显示，本文所提算法的偏好度量结果是准确的，并且很好地考虑了用户的个性化和潜在需求，而不是一味地追求结果的可靠性。

总的来说，本文提出的规则集聚合算法高效且准确，在不需要监督信息的情况下能很好地滤去冗余规则，且仅产生极小的信息损失。偏好度量框架具有良好的可扩展性，能使用任意的计算公式，极大地拓展了其在实际中的应用领域，并且在准确率、召回率和 F1-Measure 等指标上相比最新的算法也有较大的提升。

在未来工作中，为了进一步提升算法的可扩展性和准确性，将致力于以下方面的研究：

（1）当偏好数据库中用户数较少时，共识偏好较不稳定，可尝试找到一种能从较少的用户偏好中发现泛化的共识偏好的算法。

（2）引入更多种信任度函数，并通过机器学习的方式使用不同种特征的数据集进行训练，以此来给出不同情况下的最优解决方案。

（3）在考虑正偏好的同时考虑负偏好的影响，以进一步提升度量框架的准确性。

## 参 考 文 献

[1]　ZHU H S，YU K F，CAO H H，et al. Mining personal context aware preferences for mobile users[C]// Proceedings of the IEEE 12th International Conference on Data Mining. IEEE

Computer Society, 2012: 1212-1217.

[2] DE AMO S, DIALLO M S, DIOP C T, et al. Contextual preference mining for user profile construction[J]. Information Systems, 2015, 49: 182-199.

[3] WANG L C, MENG X W, ZHANG Y J. Context-Aware Recommender Systems[J]. Journal of Software, 2012, 23(1): 1-20.

[4] TAN Z, LIU J L, YU H. Conditional preference mining based on Max Clique[J]. Journal of Computer Applications, 2017, 37(11): 3107-3114.

[5] GENG L, HAMILTON H J. Interestingness measures for data mining: A survey[J]. ACM Computing Surveys (CSUR), 2006, 38(3): 9.

[6] SHIMONY S E. On Graphical Modeling of Preference and Importance[J]. Journal of Artificial Intelligence Research, 2006, 25(1): 389-424.

[7] PHILIPPE F V, VINCENT S T. Mining Top-K Non-redundant Association Rules[M]// Foundations of Intelligent Systems. Springer Berlin Heidelberg, 2012: 31-40.

[8] LAN P P, PHAN N Q, NGUYEN K M, et al. Interestingnesslab: A Framework for Developing and Using Objective Interestingness Measures[C]// International Conference on Advances in Information & Communication Technology. 2016.

[9] SZPEKTOR I. Contextual preferences[C]// Meeting of the Association for Computational Linguistics. 2008: 683-691.

[10] STEFANIDIS K, PITOURA E, VASSILIADIS P. Managing contextual preferences[J]. Information Systems, 2011, 36(8): 1158-1180.

[11] LIU J, SUI C, DENG D, et al. Representing conditional preference by boosted regression trees for recommendation[M]. Elsevier Science Inc., 2016.

[12] ZHENG L, ZHU F X, YAO X, et al. Recommendation Rating Prediction Based on Attribute Boosting with Partial Sampling [J]. Chinese Journal of Computers, 2016, 39(8): 1501-1514.

[13] EITER T, FINK M, WEINZIERL A. Preference-based inconsistency assessment in multi-context systems[C]// European Conference on Logics in Artificial Intelligence. Springer-Verlag, 2010.

[14] ZHANG X, WANG Y Q. Correlation-based interesting association rules mining[D]. Wuhan: Wuhan University of Science and Technology, 2002.

[15] TAN X Q. The Comparative Study on Interestingness Measures for Mining Association Rules[J]. Journal of the China Society for Scientific and Technical Information, 2007, 26(2): 266-270.

[16] YU C H, LIU X J, LI B. Implicit Feedback Personalized Recommendation Model Fusing Context-aware and Social Network Process[J]. Computer Science, 2016, 43(6): 248-253.

[17] LIU N N, ZHAO M, YANG Q. Probabilistic latent preference analysis for collaborative filtering[C]// Proceedings of the 18th ACM Conference on Information and Knowledge Management. Hong Kong, China: ACM, 2009.

[18] PAN W, CHEN L. GBPR: group preference based Bayesian personalized ranking for one-class collaborative filtering[C]// Proceedings of the Twenty-Third International Joint Conference on Artificial Intelligence. 2013.

[19] KYUNGYONG J, JUNGHYUN L. User preference mining through hybrid collaborative filtering and content-based filtering in recommendation system[J]. Ieice Transactions on Information & Systems, 2004, 87(12): 2781-2790.

[20] SHAHJALAL M A, AHMAD Z, AREFIN M S, et al. A user rating based collaborative filtering approach to predict movie preferences[C]// 2017 3rd International Conference on Electrical Information and Communication Technology (EICT). IEEE, 2017.

[21] GAO S, GUO G B, LIN Y S, et al. Pairwise Preference Over Mixed-Type Item-Sets Based Bayesian Personalized Ranking for Collaborative Filtering[C]// 2017 IEEE 15th Intl Conf on Dependable, Autonomic and Secure Computing, 15th Intl Conf on Pervasive Intelligence and Computing, 3rd Intl Conf on Big Data Intelligence and Computing and Cyber Science and Technology Congress(DASC/PiCom/DataCom/CyberSciTech). IEEE, 2017.

[22] HA V, HADDAWY P. Similarity of personal preferences: theoretical foundations and empirical analysis[J]. Artificial Intelligence, 2003, 146(2): 149-173.

[23] GONG Z W, Forrest J, YANG Y J. The optimal group consensus models for 2-tuple linguistic preference relations[J]. Knowledge-Based Systems, 2013, 37: 427-437.

[24] SNEDECOR G W, COCHRAN W G. Statistical Methods(Sixth-Edition)[M]. The Iowa State University Press, 1967.

[25] BOUTILIER C, BRAFMAN R I, HOOS H H, et al. Reasoning with conditional ceteris paribus preference statements[C]// Proceedings of the Fifteenth Conference on Uncertainty in Artificial Intelligence. Morgan Kaufmann Publishers Inc., 1999: 71-80.

[26] SILBERSCHATZ A, TUZHILIN A. What makes patterns interesting in knowledge discovery systems[J]. IEEE Transactions on Knowledge and data engineering, 1996, 8(6): 970-974.

[27] GRAHAM I. Non-standard logics for automated reasoning[J]. Fuzzy Sets & Systems, 1990, 36(3): 405-406.

[28] LENAT D B, GUHA R V. Building large knowledge-based systems[M]. Addison-Wesley Pub. Co, 1990.

[29] TSENG V S, WU C W, FOURNIER-VIGER P, et al. Efficient algorithms for mining top-k high utility itemsets[J]. IEEE Transactions on Knowledge and data engineering, 2015, 28(1): 54-67.

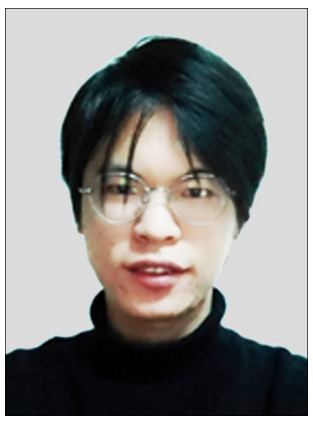

**YU Hang**, born in 1998. His main research interests include data mining, collaborative filtering, human robot interaction.

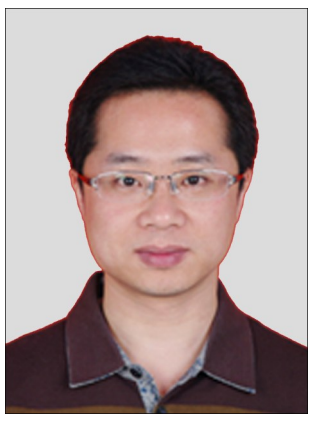

**TAN Zheng**, born in 1968, master, associate professor. His main research interests include data mining, nature language process.